\documentclass[lettersize,journal]{IEEEtran}
\usepackage{amsmath,amsfonts}
\usepackage{algorithmic}
\usepackage{algorithm}
\usepackage{array}
\usepackage[caption=false,font=normalsize,labelfont=sf,textfont=sf]{subfig}
\usepackage{textcomp}
\usepackage{stfloats}
\usepackage{url}
\usepackage{verbatim}
\usepackage{graphicx}
\usepackage{cite}
\usepackage{tabularx}
\usepackage{multirow}

\newtheorem{definition}{\textbf{Definition}}

\begin{document}

\title{Physics-Informed High-order Graph Dynamics Identification for Predicting Complex Networks Long-term Dynamics}

\author{Bicheng Wang, Junping Wang, Yibo Xue
\thanks{This work was supported in part by the National Key Research and Development Program of China under Grant 2022YFF0903300; and in part by National Natural Science Foundation of China under Grant 92167109.}
\thanks{Bicheng Wang, Junping Wang, and Yibo Xue are with the State Key Laboratory of Multimodal Artificial Intelligence Systems, Institute of Automation, Chinese Academy of Sciences, Beijing 100190, China, and also with the School of Artificial Intelligence, University of Chinese Academy of Sciences, Beijing 100049, China (e-mail: wangbicheng2022@ia.ac.cn, junping.wang@ia.ac.cn, and xueyibo2023@ia.ac.cn).}
}

\markboth{Journal of \LaTeX\ Class Files,~Vol.~14, No.~8, August~2021}%
{Shell \MakeLowercase{\textit{et al.}}: A Sample Article Using IEEEtran.cls for IEEE Journals}


\maketitle

\begin{abstract}
Learning complex network dynamics is fundamental to understanding, modelling and controlling real-world complex systems. There are two main problems in the task of predicting the dynamic evolution of complex networks: on the one hand, existing methods usually use simple graphs to describe the relationships in complex networks; however, this approach can only capture pairwise relationships, while there may be rich non-pairwise structured relationships in the network. First-order GNNs have difficulty in capturing dynamic non-pairwise relationships. On the other hand, theoretical prediction models lack accuracy and data-driven prediction models lack interpretability. To address the above problems, this paper proposes a higher-order network dynamics identification method for long-term dynamic prediction of complex networks. Firstly, to address the problem that traditional graph machine learning can only deal with pairwise relations, dynamic hypergraph learning is introduced to capture the higher-order non-pairwise relations among complex networks and improve the accuracy of complex network modelling. Then, a dual-driven dynamic prediction module for physical data is proposed. The Koopman operator theory is introduced to transform the nonlinear dynamical differential equations for the dynamic evolution of complex networks into linear systems for solving. Meanwhile, the physical information neural differential equation method is utilised to ensure that the dynamic evolution conforms to the physical laws. The dual-drive dynamic prediction module ensures both accuracy and interpretability of the prediction. Validated on public datasets and self-built industrial chain network datasets, the experimental results show that the method in this paper has good prediction accuracy and long-term prediction performance.

\end{abstract}

\begin{IEEEkeywords}
Physics-informed Machine Learning, Complex Network, System Dynamics, Graph Neural Networks.
\end{IEEEkeywords}

\section{Introduction}

The evolutionary behavior of numerous real-world complex networks, such as brains, social networks, supply networks, etc., can be modeled as dynamics on complex networks, where components inside the system are treated as nodes in the network and coupled interactions between components are treated as edges. Learning the dynamics of these complex networks facilitates the analysis and application of complex networks, including understanding the intrinsic resilience of networks \cite{liu2024deep} and predicting their future states \cite{li2024predicting}.

Current research has proposed a series of deep neural network methods to model dynamic complex interactive systems \cite{kipf2018neural}\cite{veličković2018graph}. Typically, they use a Graph Neural Network (GNN) to learn a node representation for each timestamp and predict its future trend. However, these discrete models cannot handle irregularly sampled observations and require observations at each node to be accessible at each timestamp. In contrast, the Divine Regular differential equation (ODE) approach is effective for modeling system dynamics with missing data \cite{chen2018neural}. Recent work \cite{jin2022multivariate}\cite{choi2022graph} has extended this technique to modeling interacting dynamical systems. In general, these approaches often combine GNNS with neural ODE models to capture spatio-temporal relationships in dynamic systems. Physical Information Neural Network (PINN) \cite{raissi2018deep} is proposed to solve the forward and inverse problems of partial differential equations, which provides the possibility to solve nonlinear dynamical systems. Many current studies focus on solving spatio-temporal sequence problems with PINNs. For example, PINNsFormer\cite{zhao2024pinnsformer} is based on the Transformer framework, which enables PINNs to have the ability to capture temporal dependencies through generated pseudo-sequences. Thus, the ability to solve differential equations and the approximation accuracy are enhanced. Combining tranformers with PINNs, PhysicsSolver module based on Transformer with physical attention module is proposed for spatiotemporal PDE solving in PhysicsSolver\cite{zhu2025physicssolver}.

\begin{figure*}[t]
\centering
\subfloat[]
{\includegraphics[width=.9\columnwidth]{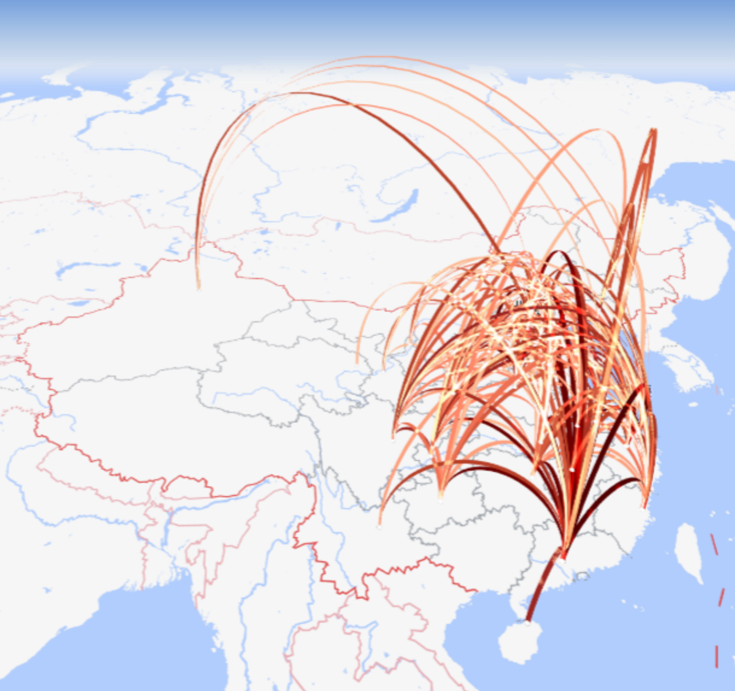}}
\centering
\hfill
\subfloat[]
{\includegraphics[width=.9\columnwidth]{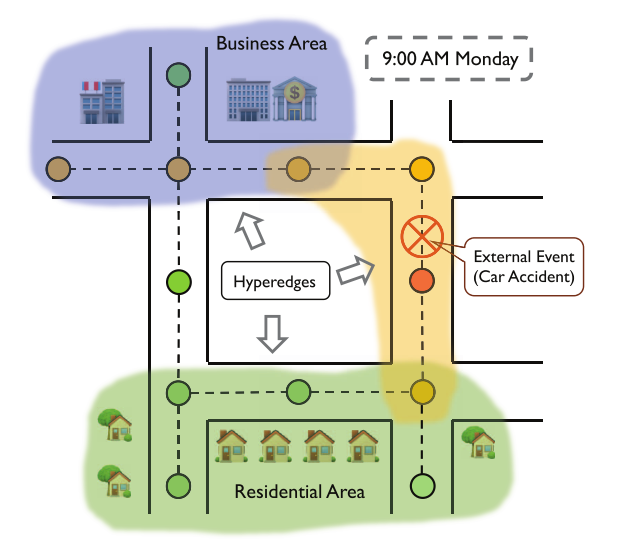}}\label{gbb}
\centering

\centering
\caption{Examples of dynamic higher-order graph structures in supply chains and transport networks. (a) In a supply chain network, the common supply relationship of multiple parties constitutes an unpaired relationship between firms \cite{li2023learning}. (b) In a transport network, both commercial and residential areas may imply static hyperedges, while external events may bring about dynamic hyperedges \cite{zhao2023dynamic}. }
\label{exp}
\end{figure*}

However, there are still two problems in the current study: on the one hand, it is unable to capture the dynamic non-pairwise relationship. Existing methods usually use GNNS to describe relationships in dynamic complex networks, which can only capture pairwise relationships. However, there may be rich non-pairwise structural relationships in the system, such as the cooperation of multiple enterprises in the supply chain network, and the road network in the transportation network, as shown in FIG. \ref{exp}. The high-order non-pairwise interaction characteristics and nonlinear dynamic nature of complex networks pose a great challenge to the existing methods. The traditional graph structure can only describe the binary relationship between nodes, and it is difficult to represent the hyperedge association formed by multi-agent synergy. For example, scenarios such as multi-firm joint production in a supply chain and group information dissemination in a social network involve high-order interactions, and it is difficult for first-order GNNS to capture such patterns only through neighborhood aggregation. Although hypergraph theory provides mathematical tools for modeling high-order relationships, the existing methods mostly rely on predefined hyperedge structures, which cannot adapt to the evolution characteristics of dynamic networks.

On the other hand, data-driven nonlinear dynamics modeling based on deep learning has powerful fitting ability, but limited by black-box characteristics and error accumulation effect, it is difficult to meet the stability and interpretability requirements of long-term prediction, and its modeling ability for high-order interactions of complex networks is limited. Complex networks are often approximated by nonlinear dynamics, and although they are feasible from a computational perspective and can be simulated using deep learning, there are currently few general frameworks for solving nonlinear dynamical systems. Therefore, representing nonlinear dynamics in a linear framework is particularly attractive because linear systems have powerful and comprehensive analysis and control techniques that do not easily generalize to nonlinear systems. In addition, the dynamic evolution of network topology in real scenarios is often accompanied by the constraints of physical laws. Due to the lack of explicit embedding of physical mechanism, the prediction results of existing methods are easy to deviate from the evolution trajectory of the actual system.

To address these issues, this chapter proposes the Physical-Informed High-orderNetworks Dynamics Identification (PhyHSL) framework, which fuses physical laws with data-driven high-order graph structure modeling. To construct a collaborative optimization paradigm for long-term dynamic prediction of complex networks. Specifically, firstly, a Dynamic Hypergraph Structure Learning (DHSL) module is introduced to break through the limitation of traditional hypergraph dependence on predefined structure, and dynamically generate adaptive hyperedges through low-rank matrix factorization and hypergraph convolution. Online modeling of non-pairwise interaction was realized. This module combines the node state representation with the learnable weight matrix, and captures the multi-scale group behavior pattern in complex networks by iteratively updating the hyperedge embedding and node embedding. Secondly, the Koopman operator theory is introduced to map the nonlinear dynamic system to the infinite dimensional linear space, and the global linearization representation is constructed. The theory approximates the original system dynamics by linear evolution in the observable function space, which not only provides a rigorous mathematical framework for nonlinear problems, but also significantly improves the stability of the model in long-term prediction. Then, a physical data-driven modeling strategy was proposed, which combined physical information Neural Ordinary Differential equations (Neural Odes) with the data-driven Koopman linearization method. The former constrains the continuous evolution trajectory of node states through differential equations to ensure that the prediction results conform to physical laws. The latter utilizes the analytical advantages of linear systems and reduces the error accumulation effect. The two models are optimized collaboratively through the variational inference framework, which enhances the robustness of the model to noisy data while preserving physical consistency. In summary, the contributions of this chapter are as follows:

\begin{itemize}
\item We introduce a dynamic hypergraph structure learning module. This module utilizes hyperedge information that exploits node associations to update node representations, thus modeling dynamic non-pairwise relationships and capturing more complex relationships in the network.
\item We propose a dual-driven dynamic prediction module for physical data. The Koopman operator theory is introduced to transform the nonlinear dynamic differential equation of complex network dynamic evolution into a linear system for solving. At the same time, the physical information neural ODE method is used to ensure that the dynamic evolution conforms to the physical law, and the error generated in the process of collecting real data through information means can be corrected, and the complex relationship between complex network subjects is more clear. The dual-drive dynamic prediction module ensures the accuracy and interpretability of the prediction.
\item is verified in the public data set and the self-built supply chain network data set. The experimental results show that our method has better prediction accuracy and generalization performance. In the face of self-built data sets with low data quality, it can also achieve high accuracy, which confirms the practical value of the method in real industrial scenarios.
\end{itemize}

The rest of this paper is organized as follows:
Section \ref{related2} presents related work on network dynamics prediction and hypergraph neural networks. Section \ref{method2} describes in detail the proposed method for predicting the long-term dynamics of complex networks by physical information high-order graph structure learning, and elaborates on the specific design of each module. In Section \ref{expre2}, our method is verified by specific experiments and analysis.
Finally, we summarize the proposed method in Section \ref{conclu2}.

\section{Related Works}
\label{related2}

\subsection{Network dynamic prediction}

The dynamic processes of real-world complex networks are nonlinear and multi-scale, and are usually abstracted into complex network models, thus illustrating the interactions between nodes. With the development of deep learning techniques such as graph neural networks, data-driven modeling of complex network dynamics has received extensive attention. Murphy et al. \cite{murphy2021deep} propose a GNN architecture that can accurately model disease spread on a network with minimal dynamical assumptions. NCDN\cite{zang2020neural} is the first to combine neural Odes and GNNs to model the continuous-time dynamics of complex networks. Huang et al \cite{huang2023generalizing} successfully model dynamic topology and cross-environment network dynamics by introducing edge dynamic Odes and environment encoders, respectively. MTGODE\cite{jin2022multivariate} proposed by Jin et al abstracts multivariate time series into a dynamic graph with time evolution node characteristics, and models its continuous dynamics in the latent space. Li et al \cite{li2024predicting} identify the skeleton of complex networks based on the renormalized group structure in hyperbolic space, which is used to predict complex network dynamics. However, most of these methods are based on first-order GNNS, and complex networks often have high-order non-pairwise relationships, so it is necessary to introduce high-order graph representations such as hypergraphs to model complex networks.

\subsection{Hypergraph Neural Networks}

A hypergraph is a generalized form of a graph consisting of a set of nodes and hyperedges. Unlike graph structured data, hypergraphs can describe non-pairwise connections because each hyperedge can be linked to many nodes. Hypergraph Neural Network (HGNN)\cite{feng2019hypergraph} is the first spatial method for hypergraph learning, which can discover potential node representations by studying higher-order structural information. However, most of these works focus on static hypergraphs. There have been recent efforts to learn dynamic hypergraphs to address this problem. Dynamic Hypergraph Neural Network (DHGNN)\cite{jiang2019dynamic} is the first effort to address hyperedge dynamics, which uses kNN and K-Means algorithms to cluster node features to build a dynamic hypergraph and iteratively perform hypergraph convolution. HGC-RNN\cite{yi2020hypergraph} uses hypergraph convolution in combination with RNN for dynamic sequence prediction. MSHyper\cite{shang2024mshyper} introduces multi-scale hypergraphs to model higher-order pattern interactions for long-term time series prediction. However, on the one hand, these methods need to rely on predefined hypergraph structure or node state similarity. On the other hand, the traditional data-driven methods are inaccurate and lack of interpretability for the dynamic modeling of complex networks. Therefore, we propose physical information high-order graph structure learning to predict the long-term dynamics of complex networks.

\section{Methodology}
\label{method2}

\begin{figure*}
    \centering
    \includegraphics[width=\linewidth]{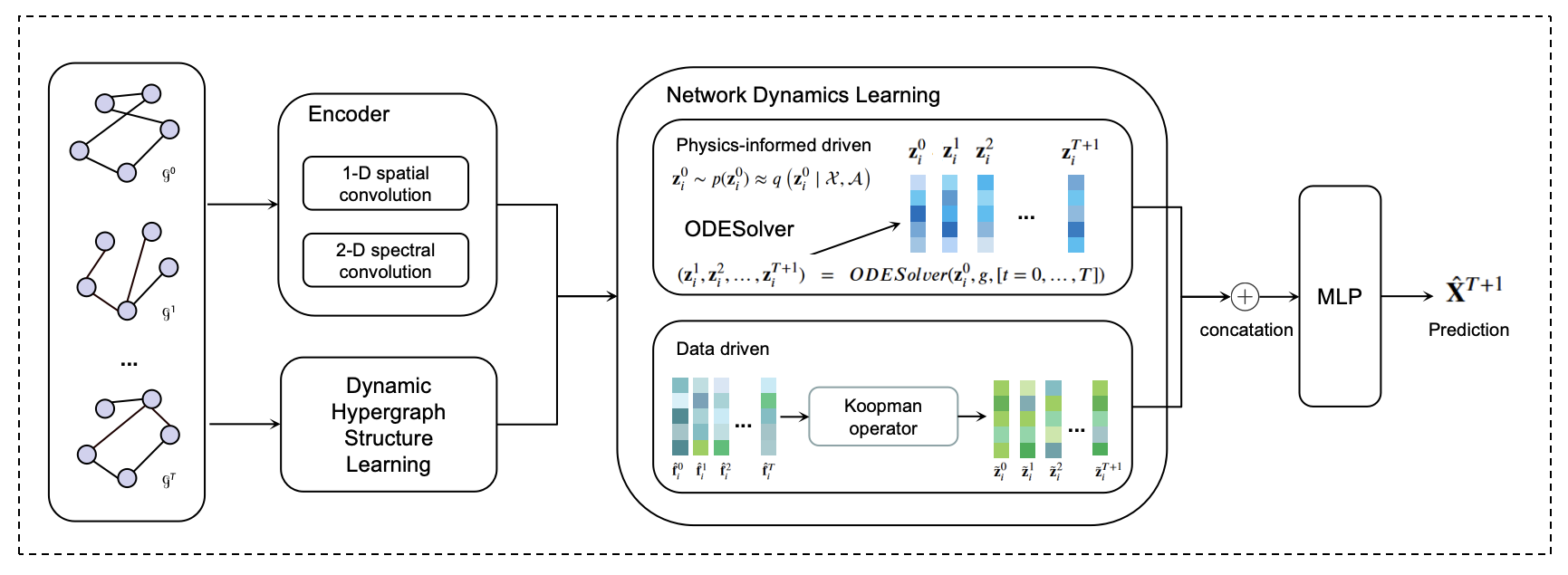}
    \caption{The framework of PhyHSL. (a) Firstly, we model the complex dynamic network as a temporal graph, and then capture the high-order adjacency relationship between nodes by first-order spatial convolution and second-order spectral domain convolution. Then, in order to study the high-order non-pairwise interaction between complex network nodes, dynamic hypergraph structure learning is introduced to generate each hyperedge embedding by aggregating the information of all connected nodes, and then the hypergraph embedding is used to update the node embedding to achieve high-order correlation learning of complex networks. Then we propose the physical information data dual driving strategy for network dynamic learning. In the physical information driven module, we introduce the nonlinear dynamic differential equation of information diffusion in the graph, and use the ODE solver to obtain the future state. In the data-driven module, we introduce the Koopman operator to transform the nonlinear dynamics into linear dynamics, so as to predict the future state. Finally, the two parts were fused and the final predicted network dynamics was obtained through the two-layer MLP.}
    \label{phyhsl}
\end{figure*}

This section will first introduce the main structure of PhyHSL, then elaborate on the proposed modules respectively, and finally introduce the loss function adopted in the training of this model. Fig. \ref{phyhsl} shows the main framework of our method.

\subsection{Domain Relation Capture Encoder}

To describe both temporal and spatial correlations, we introduce a temporal graph where each node represents an observation of an object at a given timestamp. Our graph contains two types of edges, namely, spatial and temporal edges. The spatial edge is the weighted edge between two objects based on the same timestamp, while the temporal edge is the weighted edge between every two consecutive observations of each object.

Specifically, the observation $i^t$of node $i$at time $t$in the constructed temporal graph $\mathcal{G}$. The adjacency matrix $\mathbf{A}$contains the following spatial and temporal edges:

\begin{eqnarray}
\mathbf{A}(i^t,j^{t'})  =  \begin{cases}
 w_{ij}^t &  t'  =  t\\
 1 & i  = j,t'  =  t+1 \\
  0& otherwise
\end{cases}
\end{eqnarray}

First, the domain relation capture encoder aims to learn complex spatio-temporal correlations to initialize the latent state representations of objects and edges. On the one hand, spatial convolution is combined with the attention mechanism mechanism to adaptively learn neighborhood information in the temporal graph. On the other hand, the second-order spectral graph convolution is used to explore the non-neighborhood semantic information in the embedded graph spectrum in a low-parameter manner.

\textbf{First-order spatial convolution. } In general, space-based GNNS leverage the message passing paradigm to provide discriminative node embeddings. Specifically, at each layer, they embed node semantic features into the deep representation by extracting information from the first-order neighborhood of each node. To adaptively infer the interaction between each node and its neighbors, we utilize an attention mechanism during the convolution process.

Specifically, the interaction scores between each central node and its neighbors are first computed, and these scores are used to aggregate the embeddings of its neighbors in the previous layer. Formally, a given node $I ^ t $in the first $k $l ayer embedded $\mathbf {h} ^ {t, (k)} _i $, among them, $\mathbf{h}_i^{t,(0)}=\mathbf{x}_i^t$, then its interaction score with its neighbor $j^{t'}$in the temporal graph is derived from the adjacency matrix and its features:

\begin{eqnarray}
     s ^ { ( k ) } ({ i } ^ { t } ,  { j } ^ { t ^ { \prime } } ) = \mathbf{ A } ( i _ { t } , j _ { t ^ { \prime } } ) \cos ( W _ { quer y } \mathbf{h} _ { i } ^ { t , ( k ) } , W _ { k e y } \mathbf{h} _ { j } ^ { t ^ { \prime } , ( k ) } ) 
\end{eqnarray}
where $cos(\cdot,\cdot)$ computes the cosine similarity between two vectors. The $\mathbf{W}_{query}$ and $\mathbf{W}_{key}$ are the two matrices used for similarity calculation. Then the updated $k + 1$th layer nodes are represented as follows:

\begin{eqnarray}
    \mathbf{h} _ { i } ^ { t , ( k + 1 ) } = \mathbf{h} _ { i } ^ { t , ( k ) } + \sigma ( \sum _  { j  ^ { t ^ { \prime } } \in N _ {{ i } ^ { t } } } s ^ { ( k ) } ({ i } ^ { t } , { j } ^ { t ^ { \prime } } ) \mathbf{W} _ { v a l u e } \mathbf{h} _ { j } ^ { t ^ { \prime } , ( k ) } )
\end{eqnarray}
where $\sigma$ is a nonlinear activation function and $\mathbf{W}_{value}$ is a learnable transformation matrix. The $N_{i^t}$ sets the first order neighbourhood of $i_t$. After stacking $K$ layers, a representation of each node, $\mathbf{h}_i^t$, can be obtained adaptively.

\textbf{Second-order spectral convolution. } However, spatial graph convolutions can be ineffective in exploring non-neighborhood correlations. To this end, spectral graph convolution is further introduced as a supplement to explore the semantic information hidden in the spectral domain.

Firstly, the Chebyshev polynomials are defined recursively $ T _ { 0 } ( x )=1 , T _ { 1 } ( x )=x,  T _ { m } ( x ) = 2 x T _ { m - 1 } ( x ) - T _ { m - 2 } ( x ) $. All node features are stacked into a matrix $ \mathbf{C} ^ { ( 0 ) } = \mathbf{X}$, given the normalised graph Laplacian $ \mathbf{L} = \mathbf{I} - \tilde { \mathbf{D} } ^ { - \frac { 1 } { 2 } } \mathbf{A} \tilde { \mathbf{D} } ^ { - \frac { 1 } { 2 } }$, the degree matrix $\tilde { \mathbf{D} }$ and the unit matrix $\mathbf{I}$, using second-order Chebyshev graph convolution, and the node representation matrix of the $k$th layer $\mathbf{C}^{(k)}$ is denoted as:

\begin{eqnarray}
    \mathbf{C} ^ { ( k ) } = \sum _ { m = 0 } ^ { 2 } T _ { m } ( \tilde { \mathbf{L} } ) \mathbf{C} ^ { ( k - 1 ) } \mathbf{W} _ { m } ^ { ( k ) }
\end{eqnarray}
where $ \tilde { \mathbf{L} } = 2 \mathbf{L} / \lambda _ { \ \ \ \max } - \mathbf{I}$, and $ \lambda_{max}$ denotes the maximal eigenvalue of $ \mathbf{L}$. $ \mathbf{W} _ { m } ^ { ( k ) } \in \mathbb{R} ^ { d \times d }$ is the learnable matrix of the $k$th layer. The $K$th layer is stacked and each representation vector $\mathbf{c}_i^t$ is generated from $ \mathbf{C} ^ { ( K ) } \in \mathbb{R} ^ { N \times T \times d }$ using semantics from a spectral perspective.

To generate a sequence representation for initialization, the two representations are first combined and then these temporal representations are aggregated into a sequence representation for each object using an attention mechanism.
Specifically, the two representations of the last layer are first combined with the temporal embedding, and then MLP is utilized to obtain the final representation of each $i^t$. In form:

\begin{eqnarray}
     \mathbf{q} _ { i } ^ { t } = MLP ( \left[ \mathbf{c} _ { i } ^ { t } , \mathbf{h} _ { i } ^ { t } \right] ) + T E ( t )\end{eqnarray}
     \begin{eqnarray}
     T E ( t ) [ 2 i ] = \sin ( \frac { \Delta t } { 1 0 0 0 0 ^ { 2 i / d } } )\end{eqnarray}
     \begin{eqnarray}
    T E ( t ) [ 2 i + 1 ] = \cos \left( \frac { \Delta t } { 1 0 0 0 0 ^ { 2 i / d } } \right) 
\end{eqnarray}
where $[2i]$and $[2i+1]$ represent the element index of the parity position in the temporal embedding, respectively. Then, using the attention operator, these node representations at each time are summarized into a summary representation $\mathbf{u}_i$, which is denoted as follows:

\begin{eqnarray}
     \mathbf{u} _ { i } ^ { t } = \frac { 1 } { T } \sum _ { t = 1 } ^ { T } \sigma (\mathbf{v}_{q_1} \alpha _ { i } ^ { t } \mathbf{q} _ { i } ^ { t } ) 
\end{eqnarray}
\begin{eqnarray}
     \alpha _ { i } ^ { t } = \frac{\mathrm{exp}(\sigma (\mathbf{v}_{q_2} \alpha _ { i } ^ { t } \mathbf{q} _ { i } ^ { t } ) )}{\sum _ { t = 1 } ^ { T }\mathrm{exp}(\sigma (\mathbf{v}_{q_2} \alpha _ { i } ^ { t } \mathbf{q} _ { i } ^ { t } )  ) }
\end{eqnarray}
\begin{eqnarray}
    \mathbf{u}_i \in \mathbb{R}^{T \times d} =\{ \mathbf{u}_i^t\}_{t=1}^T 
\end{eqnarray}
where $\alpha _ { i } ^ { t }$ denotes the normalised attention score and $\mathbf{v}_{q_1},\mathbf{v}_{q_2}$ are learnable parameters.

\subsection{Dynamic Hypergraph Structure Learning}

\begin{figure}[t]
    \centering
    \includegraphics[width=\linewidth]{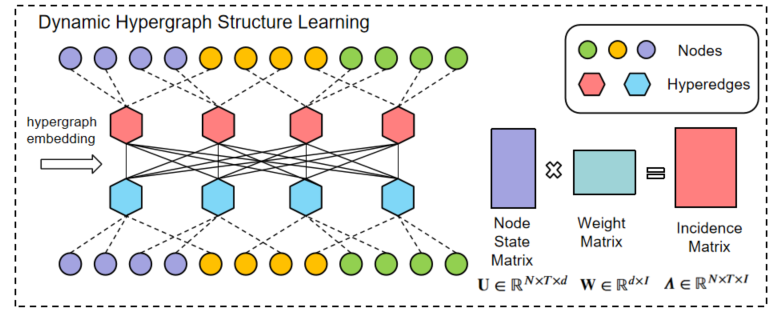}
    \caption{Dynamic hypergraph structure learning module.We first generate a low-rank correlation matrix using the node features. The hypergraph convolution operation first fuses the information from the connected nodes into the hyperedge representation and then reconstructs the node representation using the associated hyperedge representation.}
    \label{hyper}
\end{figure}

There are complex interactions between nodes in complex networks, and in many cases they do not exist in pairs, so it is necessary to model the higher-order structure of the network. However, most of the previous methods are based on dynamic graphs or dynamic hypergraphs, which need to predefine the network structure \cite{yi2020hypergraph}. There are often problems such as high computational cost \cite{guo2019attention} and inability to capture higher-order non-pairwise relationships \cite{song2020spatial}. Therefore, referring to the work of Zhao et al \cite{zhao2023dynamic}, we introduce the dynamic hypergraph structure learning module, as shown in Figure \ref{hyper}.

First, we formalize the incidence matrix of the temporal hypergraph as $\mathbf{\Lambda}\in \mathbb{R}^{N\times T \times I}$, where $I$is the number of the hyperedges. The matrix is decomposed into two low-rank matrices using their hidden states as follows:

\begin{eqnarray}
    \mathbf{\Lambda} = \mathbf{UW}_{\Lambda}
\end{eqnarray}
where $\mathbf{U}\in \mathbb{R}^{N\times T\times d}$ is obtained by superposition of all state representations, and $\mathbf{W}\in \mathbb{R}^{d×I}$ is the learnable weight matrix.

Then, a hypergraph convolution paradigm is introduced to learn temporal hypergraphs, which can extract higher-order complex information from dynamic complex networks. Specifically, at each layer, each hyperedge embedding is first generated by aggregating information from all connected nodes, and then the node embeddings are updated using the hypergraph embeddings to achieve higher-order relevance learning of complex networks. The whole process is summarised in Fig. \ref{hyper}

In matrix form, the hyperedge embedding matrix $\mathbf{E}\in \mathbb{R}^{I\times d}$ is derived from the state representation matrix and the association matrix:.

\begin{eqnarray}
    \mathbf{E} = \sigma ( \mathbf{W}_E \mathbf{\Lambda ^ { T } U} ) + \mathbf{\Lambda ^ { T } U }
\end{eqnarray}
where $\mathbf{W}_E$ is a learnable matrix to represent the implicit relationship between hyperedges. Then, these hyperedge embeddings are aggregated to generate the node embedding matrix as follows:

\begin{eqnarray}
    {\mathbf{F} }_i &=&\mathbf{\Lambda E}\nonumber\\  &=&\mathbf{\Lambda}(\sigma ( \mathbf{W}_E\mathbf{\Lambda ^ { T } U} ) + \mathbf{\Lambda ^ { T } U })
\end{eqnarray}

In this way, we can learn the complex high-order graph structure of the network at each timestamp by stacking $L$hypergraph convolutional layers, and output the updated node embedding matrix ${\mathbf{F}}\in \mathbb{R}^{N\times T\times d}$ where ${\mathbf{f}}_i^t$ represents the state of node $i$ at time $t$.

\subsection{Network Dynamic Learning}\label{phydy}

\textbf{Physis-informed driven:} Firstly, we generate a state-initialised representation of the sequence model by sampling an approximate posterior distribution, i.e., $q\left(\mathbf{z}_{i}^{0} \mid \mathcal{X}, \mathcal{A}\right)$, based on the sequence representation. In addition, during the optimisation process, the posterior distribution should be approximated as regularised to the prior distribution $p(\mathbf{Z}^0)$. To achieve this, the mean and variance of the posterior distributions are measured to minimise their differences, and a ‘reparameterisation’\cite{kingma2013auto} is used from the posterior distributions when generating the initialised state vector $\mathbf{z}_i^0$ of node i, respectively. Formally:

\begin{eqnarray}
     q\left(\mathbf{z}_{i}^{0} \mid \mathcal{X}, \mathcal{A}\right)=\mathcal{N}\left(MLP^{m}\left(\mathbf{f}_{i}\right), MLP^{v}\left(\mathbf{f}_{i}\right)\right) 
\end{eqnarray}
\begin{eqnarray}
    \mathbf{z}_i^0 \sim p(\mathbf{z}_i^0) \approx q\left(\mathbf{z}_{i}^{0} \mid \mathcal{X}, \mathcal{A}\right)
\end{eqnarray}
where $\mathcal{N}$ denotes the normal distribution.

Given an initial state $\mathbf{z}_i^0$, we apply the neural ODE solver \cite{chen2018neural} to compute the future node state $\mathbf{z}^{(1:T)}$:

\begin{eqnarray}
    (\mathbf{z}_i^1,\mathbf{z}_i^2,\dots,\mathbf{z}_i^{T+1})=ODESolver(\mathbf{z}_i^0, g,[t=0,\dots,T])
\end{eqnarray}

\textbf{Data-driven:} In addition, complex networks are usually approximated by nonlinear dynamics, which, although feasible from a computational point of view and can be solved by simulation via neural ODEs, are still not comprehensively accurate in terms of their solution and analytical frameworks in comparison to linear systems.Koopman operator theory\cite{strogatz2018nonlinear}\cite{ mezic2020koopman} maps nonlinear systems to global linear systems in the space of observable functions, and is widely used in the solution and analysis of nonlinear dynamical systems\cite{mezic2024koopman}\cite{liu2023koopa}\cite{hogg2020exponentially}.

\begin{definition}
    The nonlinear temporal dynamical system can be represented as $x_{t+1} = \mathbf{F}(x_t)$, where $x_t$represents the state of the system and $\mathbf{F}$is the vector field describing the dynamics. However, it is challenging to identify the transition of the system directly on the state due to the presence of nonlinear or noisy data. In contrast, Koopman theory assumes that states can be projected into the space of measurement functions $g$, which can be governed by an infinite-dimensional linear operator $\mathcal{K}$and advanced in time such that:

\begin{eqnarray}
     \mathcal{K} \circ g ( x _ { t } ) = g ( \mathbf{F} ( x _ { t } ) ) = g ( x _ { t + 1 } ) 
\end{eqnarray}
\end{definition}

\begin{figure}[t]
    \centering
    \includegraphics[width=\linewidth]{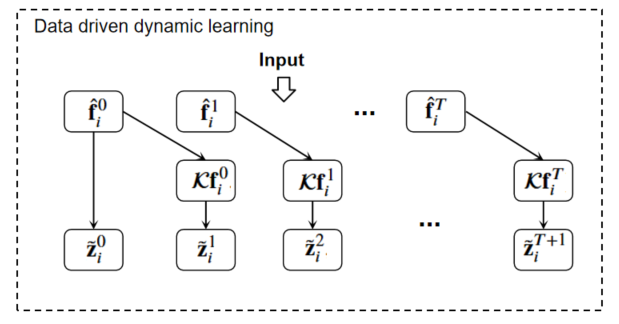}
    \caption{Data-driven Koopman operator dynamic learning module.}
    \label{koopman}
\end{figure}

As a result, the data-driven latent state representation can be obtained as follows.

\begin{eqnarray}
    (\tilde{\mathbf{z}}^0_i,\tilde{\mathbf{z}}^1_i,\tilde{\mathbf{z}}^2_i,\dots,\tilde{\mathbf{z}}^{T+1}_i)=(\mathbf{f}_i^0,\mathcal{K}\mathbf{f}_i^0,\mathcal{K}\mathbf{f}_i^1,\dots,\mathcal{K}\mathbf{f}_i^T)
\end{eqnarray}

In which, unlike traditional numerical analysis methods such as Dynamic Mode Decomposition (DMD)\cite{schmid2010dynamic} to approximate the operator, $\mathcal{K}$ can be derived by fitting through the data using deep learning, and the computational process is shown in Fig. \ref{koopman}

Finally, we concatenate the two parts and use the two MLP layers to obtain:

\begin{eqnarray}
    \hat{\mathbf{x}}_i^t=MLP(\sigma([\mathbf{z}_i^t,\tilde{\mathbf{z}}_i^t]))
\end{eqnarray}
where $[\cdot,\cdot]$ denotes the concatenation of vectors. Eventually, the state representation of the node $\hat{\mathbf{X}}^{T+1}\in \mathbb{R}^{N\times T\times d}$ is obtained.

In the implementation, each training sample is divided into two parts based on time and the first part is used to predict the second part. The framework is then optimized using the variational inference paradigm \cite{kingma2013auto}, i.e., maximizing the Evidence Lower Bound of Likelihood (ELBO) while minimizing the Kullback-Leibler (KL) scatter between the prior and posterior distributions. The loss function is denoted as:

\begin{eqnarray}
    \mathcal{L}-\sum_i\sum_t\frac{||\mathbf{x}_i^t-\hat{\mathbf{x}}_i^t||^2}{2\sigma^2}-KL \left[ \prod _ { i = 1 } ^ { N } q ( \mathbf{z} _ { i } ^ { 0 } | \mathcal{X , A} ) | | p ( \mathbf{Z} ^ { 0 } ) \right]
\end{eqnarray}
where $\sigma^2$ denotes the variance of the prior distribution.

\section{Experiments}
\label{expre2}

In this section, we construct complex networks using supply network data obtained from the real world and use it to evaluate the effectiveness of our proposed model.

\subsection{Experiments setup}

The method proposed in this chapter is experimented in several complex network datasets. We first evaluate the performance of PhyHSL on three publicly available datasets: Common likes Facebook page (Social), link to www.epa.gov (Web) \cite{li2024predicting}, synthetic Watts-Strogatz network (WS) \cite{watts1998collective}. Then experiments are carried out on three self-built supply chain network datasets to verify the generalization: Manufacture, Electronic and Finance supply chain network. The details of the network are shown in table \ref{table-data2}, and the supply chain network dataset has more and more complex link relationships compared with the public datasets. We repeat the experiment 10 times with different random partitions and report the results based on the test accuracy of these 10 runs. We implemented the model in PyTorch and performed all training and testing tasks on two NVIDIA A100 Gpus.

\begin{table*}
\renewcommand{\arraystretch}{1.5}

\caption{Details of datasets}
\label{table-data2}
\centering
\setlength{\tabcolsep}{3mm}
\begin{tabular}{c|cccccc}
\hline
& Social & Web & WS & Manufacture &Electronic&  Finance\\
\hline

Nodes &3892 & 4252 & 5000 & 960 &700 & 1500\\

Edges &17239 & 8896 & 10000 & 25142 &16604& 61218\\

\hline
\end{tabular}

\end{table*}


We have selected three categories of methods as experimental baseline models that will be evaluated by our method.

\textbf{GNN-based method:} 
\begin{itemize} 
 \item \textbf{DCRNN}\cite{li2017diffusion} uses a bidirectional random walk and an encoder-decoder architecture with timed sampling to achieve accurate temporal prediction.
    \item \textbf{MTGODE}\cite{jin2022multivariate}abstracts multivariate time series into dynamic graphs characterized by time-evolving nodes and models their continuous dynamics in potential space.
    \item \textbf{DiskNet}\cite{li2024predicting}Identifies the skeleton of complex networks based on the structure of reified groups in hyperbolic space for predicting complex network dynamics.
\end{itemize}

\textbf{Hypergraph-based methods:} 
\begin{itemize} 
 \item \textbf{HGC-RNN}\cite{yi2020hypergraph}Combining hypergraph convolution with RNN.
    \item \textbf{MSHyper}\cite{shang2024mshyper}Introducing multiscale hypergraphs to model higher-order pattern interactions for long-term time series prediction.
\end{itemize} 

\textbf{PINN-based methods:} 
\begin{itemize} 
 \item \textbf{PhyCRNet}\cite{ren2022phycrnet} defines the loss function as the residuals of the PDEs, utilizing autoregression (AR), residual linkage and ConvLSTM to learn temporal features.
    \item \textbf{PINNsFormer}\cite{zhao2024pinnsformer}Based on the Transformer framework, it gives PINNs the ability to capture temporal dependencies through the generated pseudosequences, which enhances the ability to solve differential equations and approximation accuracy.
    \item \textbf{PhysicsSolver}\cite{zhu2025physicssolver} combines Tranformer with PINN and proposes a physics solver module based on Transformer containing a physics attention module for space-time PDE solving.
\end{itemize}

\subsection{Results and analysis}

In this section, to quantify the performance of the model, we use the mean absolute error (MAE) as an evaluation metric:

\begin{eqnarray}
MAE  =\frac{1}{N}\sum_{i=1}^{N}||\mathbf{\hat{x}}_i-\mathbf{x} _i||
\end{eqnarray}
where $\mathbf{x} _i$ and $\mathbf{\hat{x}}_i$ are the reference and predicted solutions, respectively. The experimental results are shown in Tables \ref{table_pre} and \ref{table_pre1}.

\begin{table*}[h]
\renewcommand{\arraystretch}{2}
\caption{Performance comparison in open datasets.(The best result is in bold face.)}
\label{table_pre}
\centering
\setlength{\tabcolsep}{8mm}

\scalebox{1}{

\begin{tabular}{c|c|ccc}
\hline
& Datasets& Social & Web & WS  \\
\hline
\multirow{3}*{GNN} 
&DCRNN  & 0.487 $\pm $ 0.048 & 0.423 $\pm$ 0.045 & 0.344 $\pm $ 0.033  \\
&MTGODE & 0.226 $\pm $ 0.006 & 0.179 $\pm$ 0.008 & 0.129 $\pm $ 0.005 \\
&DiskNet &0.209 $\pm $ 0.005&\textbf{0.165 $\pm $ 0.002} &0.136 $\pm $ 0.003\\
\hline
\multirow{2}*{Hypergraph}
&HGC-RNN & 0.370 $\pm$ 0.018 & 0.317 $\pm$ 0.027 & 0.355 $\pm$ 0.015\\
&MSHyper & 0.213 $\pm$ 0.016 & 0.245 $\pm$ 0.026 & 0.155 $\pm$ 0.011\\
\hline
\multirow{3}*{PINN}
&PhyCRNet &0.412 $\pm$ 0.031&0.386 $\pm$  0.025 &0.391 $\pm$ 0.015 \\
&PINNsFormer   &0.304 $\pm$ 0.011&0.286 $\pm$  0.011 &0.191 $\pm$ 0.025\\
&PhysicsSolver &0.313 $\pm$ 0.011&0.236 $\pm$  0.020 &0.165 $\pm$ 0.008\\
\hline
&\textbf{PhyHGL} &\textbf{0.201 $\pm $ 0.007} &0.178 $\pm $ 0.014& \textbf{0.127 $\pm$ 0.007}\\
\hline
\end{tabular}
}
\end{table*}

\begin{table*}[h]
\renewcommand{\arraystretch}{2}
\caption{ Performance comparison in supply network datasets.}
\label{table_pre1}
\centering
\setlength{\tabcolsep}{8mm}

\scalebox{1}{

\begin{tabular}{c|c|ccc}
\hline
& Datasets&  Manufacture & Electricity& Finance \\
\hline
\multirow{3}*{GNN} &DCRNN  &0.349 $\pm$ 0.037& 0.345 $\pm$ 0.033 &  0.407 $\pm $ 0.014 \\
&MTGODE & 0.142 $\pm$ 0.013 & 0.313 $\pm$ 0.009& 0.187 $\pm $ 0.033 \\
&DiskNet & 0.131 $\pm$ 0.021& 0.257 $\pm$ 0.003&0.155 $\pm $ 0.028\\
\hline
\multirow{2}*{Hypergraph}&HGC-RNN & 0.223 $\pm$ 0.031 &0.297 $\pm$ 0.011 & 0.241 $\pm$ 0.016\\
&MSHyper & 0.120 $\pm $ 0.012 & 0.252 $\pm$ 0.023 &\textbf{0.154 $\pm$ 0.023}\\
\hline
\multirow{3}*{PINN}&PhyCRNet&0.298 $\pm$  0.016 & 0.288 $\pm$ 0.015& 0.366 $\pm$ 0.016 \\
&PINNsFormer & 0.214 $\pm$ 0.034 & 0.271 $\pm$ 0.021 &0.329 $\pm$ 0.029 \\
&PhysicsSolver &0.213 $\pm$ 0.009 & 0.271 $\pm$ 0.014  & 0.264 $\pm$ 0.019 \\
\hline
&\textbf{PhyHGL} &\textbf{0.112 $\pm $ 0.014} & \textbf{0.247 $\pm$ 0.013}&0.162 $\pm $ 0.027\\
\hline
\end{tabular}
}
\end{table*}

The PhyHGL proposed in this paper achieves optimal or sub-optimal results in all six datasets, which indicates the effectiveness and generalisation of our proposed method. Meanwhile, the small deviation of PhyHSL results in several experiments indicates the robustness of our method. Specifically, our method improves the accuracy by about 1\% compared to GNN-based models, and in addition to considering pairwise dependencies between nodes, our model also considers unpaired relationships, which is able to capture higher-order interactions in the graph, and exhibits better performance in dealing with complex network tasks. While most hypergraph-based approaches rely on predefined hypergraph structures, our approach fuses physical mechanisms and higher-order graph structure learning, and thus captures complex network dynamics more effectively. And compared with the latest PINN-based methods, our method improves about 10\%, indicating that purely physically-driven models may be limited by the ability to model the higher-order interactions of complex networks, relying more on modelling the dynamic equations, lacking a priori knowledge of the complex network, and not being able to handle better in the face of noisy data. In addition, from the dataset point of view, in more complex supply networks, our method with the hypergraph-based method is more capable of showing better performance compared to other methods. Taken together, PhyHGL fuses physical mechanism and data-driven through the theory of physically-informed neural ODEs and Koopman operators to enhance the nonlinear dynamic modelling capability, joint learning between physical-driven and data-driven reduces the error accumulation, and variational inference and KL scatter regularisation enhances the generalizability. The dynamic hypergraph convolution module effectively captures the higher-order interactions of unpaired nodes without the need of a pre-defined hypergraph structure, which makes up for the inadequacy of traditional graph models that only capture pairwise relationships in handling complex network tasks.

\subsection{Ablation Study}

We conduct ablation experiments on two datasets, Social and Manufacture, for each of the four components: (1) physical information-driven module (w/o Phy); (2) Koopman operator-driven module (w/o Koop); (3) Dynamic Hypergraph Structure Learning Module (w/o DHSL); and (4) dynamic hypergraph structure learning module only (w/o Phy $\&$ Koop), to assess the effect of these modules on the experimental results. The results are shown in Fig. \ref{ablation}, where the physics-informed driven module and the Koopman operator driven module act as complementary modules, discarding any of them will lead to performance degradation, and the presence of the two components also adds interpretability to our model. In addition, the presence or absence of the higher-order hypergraph structure learning module has an insignificant effect on model performance enhancement in the Soicial dataset, whereas the higher-order graph structure learning module has a more significant performance enhancement in the industrial chain network with more complex linking relationships.

\begin{figure*}[t]
\centering
\subfloat[]
{\includegraphics[width=\columnwidth]{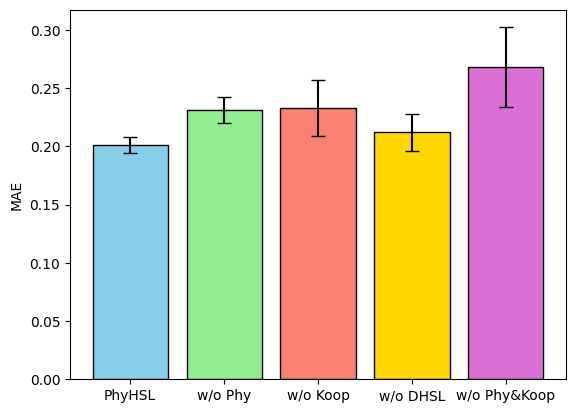}}
\centering
\hfill
\subfloat[]
{\includegraphics[width=\columnwidth]{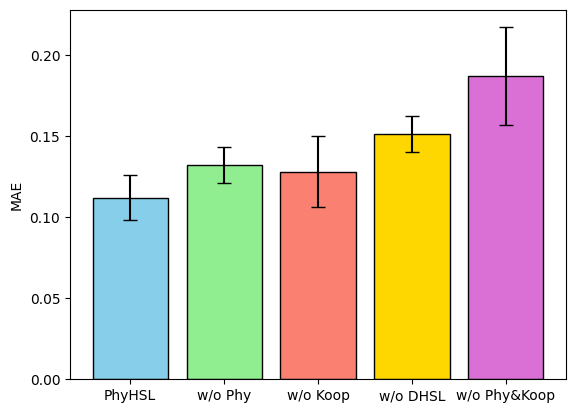}}\label{gbb}
\centering

\centering
\caption{Ablation Study. }
\label{fig1}
\end{figure*}

\begin{table}
\renewcommand{\arraystretch}{2}
\caption{Ablation Study.}
\label{table_pre1}
\centering
\setlength{\tabcolsep}{3.5mm}

\scalebox{1}{

\begin{tabular}{c|cc}
\hline
 Datasets & Social&  Manufacture\\

\hline

w/o Phy  &0.231 $\pm$ 0.011& 0.132 $\pm$ 0.011  \\

w/o Koop & 0.233 $\pm$ 0.024 & 0.128 $\pm$ 0.022 \\

w/o DHSL & 0.212 $\pm$ 0.016& 0.151 $\pm$ 0.011\\

w/o Phy\&Koop & 0.268 $\pm$0.034& 0.187 $\pm$ 0.030\\

PhyHSL & 0.201 $\pm$ 0.007 &0.112 $\pm$ 0.014\\
\hline

\end{tabular}
}

\end{table}

\subsection{Analytical Experiments} 

\begin{figure*}
\centering

\subfloat[]
{\includegraphics[width=\columnwidth]{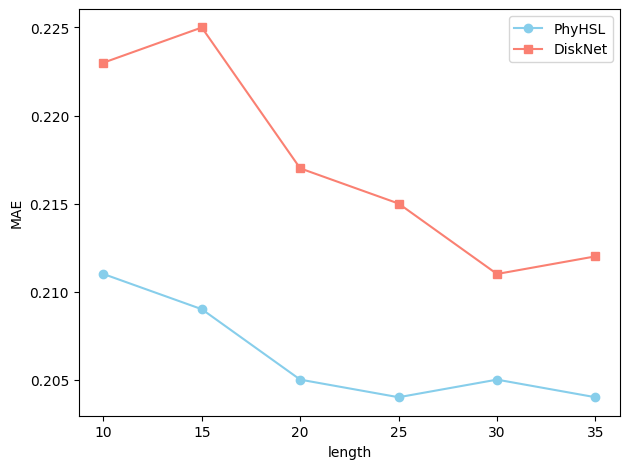}}
\centering
\hfill
\subfloat[]
{\includegraphics[width=\columnwidth]{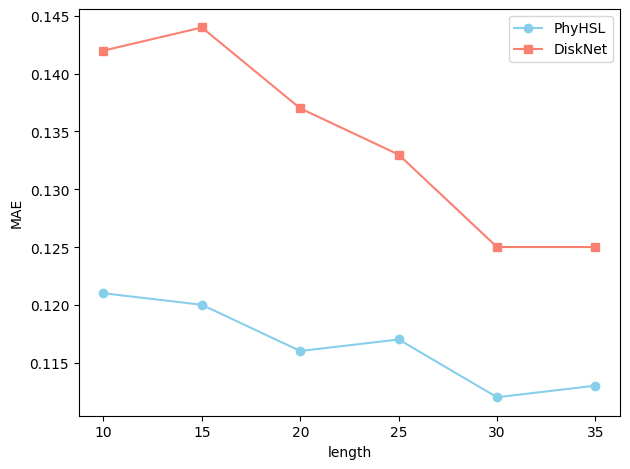}}
\centering

\centering
\caption{Model performance under different training lengths. }
\label{10len}
\end{figure*}

The prediction performance is analysed for different training lengths and prediction lengths. Firstly, we conducted experiments on the effect of different training lengths on the performance, and selected the training lengths of [10,15,20,25,30,35] on Social dataset and Manufacture respectively to test its performance in predicting the future 10 lengths, as shown in Fig. \ref{10len}. Then experiments are conducted on the model performance under different prediction lengths, and a training length of 30 is selected on Social dataset and Manufacture respectively to test its performance in predicting the future [5,8,10,12,15,20] lengths, as shown in Fig. \ref{len}. It can be seen that in most cases, our method achieves better performance and stability in terms of MAE, which validates the superiority of our method E. In addition, it can be noticed that the gap between PhyHSL and DiskNet tends to be larger in most cases when making long-term predictions. The possible reasons for this are, on the one hand, that long-term prediction relies more on higher-order dependencies, which can be captured in our method, and, on the other hand, our method adds physical information constraints, which will be less affected by error accumulation when performing long-term prediction.

\begin{figure*}\label{len}
\centering
\subfloat[]
{\includegraphics[width=\columnwidth]{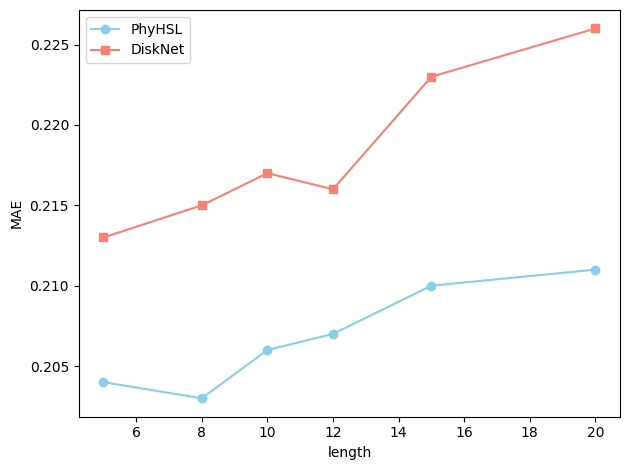}}
\centering
\hfill
\subfloat[]
{\includegraphics[width=\columnwidth]{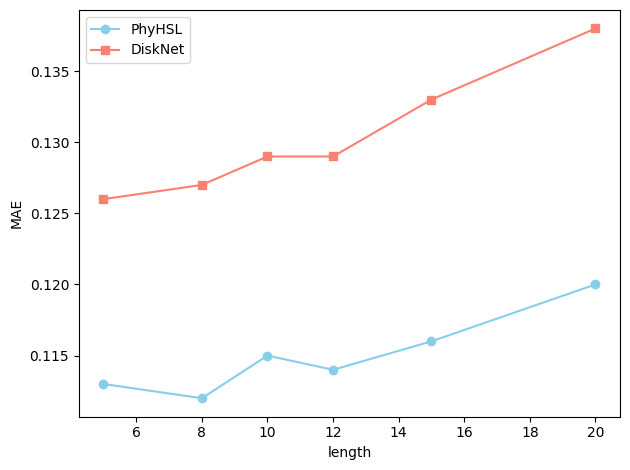}}
\centering

\centering
\caption{Model performance under different prediction lengths. }
\label{len}
\end{figure*}

\subsection{Computational efficiency analysis}

\begin{figure}
    \centering
    \includegraphics[width=\linewidth]{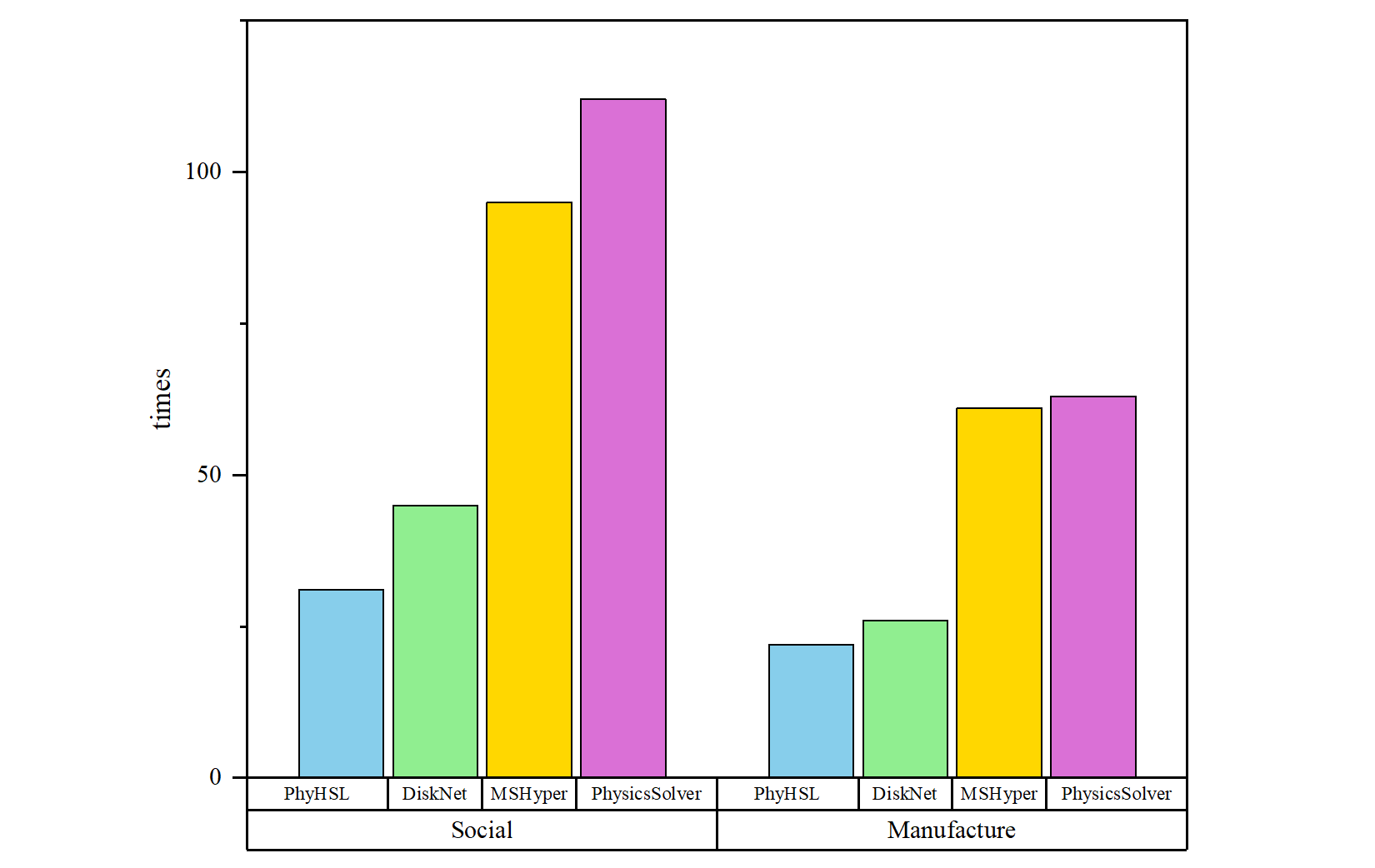}
    \caption{Time cost per iteration on different models.}
    \label{jisuan}
\end{figure}

In order to explore the efficiency of the PhyHSL model, this section compares the inference time of PhyHSL with DiskNet, MSHyper, and PhysicsSolver on the test set of both Social and Manufacture datasets on a comparative basis, as shown in Fig. \ref{jisuan}. Compared with DiskNet, the process of identifying the skeleton has a higher computational complexity, although it only requires the integration of the forward ODE function on the network skeleton, which greatly saves computational resources compared to previous neural ODE-based approaches.MSHyper and PhysicsSolver rely more on the multi-layer Transformer structure, which increases the computational consumption. In addition to this, the use of Koopman operator also reduces the set of learnable parameters, which helps to improve the computational efficiency.

\section{Conclusion}
\label{conclu2}

In this paper, we propose a physics-informed higher-order graph structure learning (PhyHGL) framework, which aims to address the core challenges of insufficient spatio-temporal feature extraction, difficulty in modeling nonlinear dynamics, and inadequate capture of higher-order interactions in the long-term dynamic prediction of complex networks. By fusing physical laws with data-driven higher-order graph structure modeling, PhyHGL constructs a collaborative optimization framework to achieve accurate modeling and prediction of complex network evolution. First, for neighborhood relationship modeling, an adaptive neighborhood capture encoder is designed, combining first-order spatial convolution and second-order spectral domain convolution to effectively balance the needs of local interaction and non-global feature extraction. In order to further enhance the physical consistency of dynamic modeling, a dual-drive strategy is proposed: on the one hand, continuous dynamics modeling based on physically informative neural ODE captures the continuous evolution trajectory of the system through a differential equation solver; on the other hand, the Koopman operator theory is introduced to map the nonlinear system into an infinite-dimensional linear space to achieve global linearized modeling, which synergistically enhances the model adaptability and interpretation to nonlinear dynamics. Aiming at the higher-order unpaired interactions existing in complex networks, we introduce the dynamic hypergraph structure learning method to dynamically update the hyperedge and node embedding, which provides the possibility of exploring the higher-order interactions of complex networks. Experimental results show that our model outperforms the benchmark model in terms of accuracy and robustness.

In the future work, we will explore the construction of more complex network relationships, such as the study of hypergraph network. In addition, the existing research is based on the known data in the past, and the real data is updated in real time. Take the supply network as an example, the supply network network will add new edges or new links to the old chain, and the network structure should be constantly updated. While traditional map neural network methods often show a significant decline in the performance of past tasks when learning new tasks, we will study more effective ways to achieve real-time monitoring and regulation of network resilience in the future

\section*{Acknowledgments}
The authors would like to express their sincere gratitude to the anonymous reviewers for their insightful comments and constructive feedback, which greatly contributed to improving the overall quality of this paper. This work was supported in part by the National Key Research and Development Program of China under Grant 2022YFF0903300; and in part by National Natural Science Foundation of China under Grant 92167109.


\bibliographystyle{IEEEtran}
\bibliography{IEEEabrv,sample}

%

\begin{IEEEbiographynophoto}{Bicheng Wang}
 is currently pursuing the M.S. degree at Institute of Automation, Chinese Academy of Sciences. He received the B.S. degree in vehicle engineering from Zhejiang University, Hangzhou, China, in 2022. 

His research interests include physics-informed machine learning, complex network, and graph data learning.
\end{IEEEbiographynophoto}

\begin{IEEEbiographynophoto}{Junping Wang}
 is currently a full professor at Institute of Automation, Chinese Academy of Sciences. He received the M.S. degree (2006) from Stanford University USA. and the PH.D degrees (2013) in computer science and technology from Beijing University of Posts and telecommunications, China. 

His research interests include pattern recognition, machine learning, and brain autonomous cognitive computer. His research has been supported by several research grants from National Science Foundation of China, and he has published more than 120 peer-reviewed papers. 
\end{IEEEbiographynophoto}

\begin{IEEEbiographynophoto}{Yibo Xue}
 is currently pursuing the M.S degree at the Institute of Automation, Chinese Academy of Sciences. He received his B.S. degree in Electronic Information from Beijing Jiaotong University in 2023. 
 
 His research interests include graph neural networks, hypergraph machine learning, and complex networks.
\end{IEEEbiographynophoto}

\vfill

\end{document}